\documentclass{article}

\usepackage[preprint]{neurips_2025}
 
\usepackage[utf8]{inputenc} % allow utf-8 input
\usepackage[T1]{fontenc}    % use 8-bit T1 fonts
\usepackage{hyperref}       % hyperlinks
\usepackage{url}            % simple URL typesetting
\usepackage{booktabs}       % professional-quality tables
\usepackage{amsfonts}       % blackboard math symbols
\usepackage{nicefrac}       % compact symbols for 1/2, etc.
\usepackage{microtype}      % microtypography
\usepackage{xcolor}         % colors
\usepackage{amsmath}
\usepackage{graphicx}
\usepackage{multirow}
\usepackage{listings}
\usepackage{tabularx}

\definecolor{cvprblue}{rgb}{0.21,0.49,0.74}
\definecolor{maroon}{RGB}{255, 20, 147}
\hypersetup{
    colorlinks=true,            %链接颜色
    linkcolor=maroon,             %内部链接
    filecolor=magenta,          %本地文档
    urlcolor=cyan,              %网址链接
    citecolor=cvprblue,             % 引用颜色
    pdftitle={Overleaf Example},
    pdfpagemode=FullScreen,
}

\title{SimuGen: Multi-modal Agentic Framework for Constructing Block Diagram-Based Simulation Models}

\author{%
Xinxing Ren$^{1,}$\thanks{Equal contribution.} \quad Qianbo Zang$^{2,*}$ \quad \textbf{Zekun Guo}$^{3,}$\thanks{Correspondence: \href{mailto:z.guo2@hull.ac.uk}{z.guo2@hull.ac.uk}.}\\
$^1$Brunel University of London  \quad $^2$SnT, Université du Luxembourg \\ $^3$University of Hull\\
}

\begin{document}

\maketitle

\begin{abstract}
Recent advances in Large Language Models (LLMs) have demonstrated remarkable capabilities in mathematical reasoning and code generation. However, LLMs still perform poorly in the simulation domain, especially when tasked with generating Simulink models, which are essential in engineering and scientific research. Our preliminary experiments reveal that LLM agents struggle to produce reliable and complete Simulink simulation codes from text-only inputs, likely due to insufficient Simulink-specific data during pre-training. To address this gap, we introduce \textbf{SimuGen}, a multi-modal agentic framework designed to automatically generate accurate Simulink simulation code by leveraging both the visual Simulink diagram image and domain knowledge. SimuGen coordinates several specialized agents—including an Investigator, a unit test reviewer, a code generator, an executor, a debug locator, and a report writer—supported by a domain-specific database. This collaborative, modular architecture enables interpretable and robust simulation generation for Simulink. Our codes are public available.~\footnote{\url{https://github.com/renxinxing123/SimuGen_beta}}
% ~\footnote{\url{https://anonymous.4open.science/r/SimuGen-CD71}}
\end{abstract}

\section{Introduction}
\textit{“One test result is worth one thousand expert opinions.”}  
\begin{flushright}
\textit{--- Wernher von Braun, Father of Rocket Science}
\end{flushright}

Simulation models are indispensable for testing complex systems across domains such as automotive, aerospace, and robotics. In industry and academia, Simulink, a tool for modeling and simulation of dynamic systems in the MATLAB environment, has emerged as a prominent block-diagram platform for simulation~\cite{mathworks_simulink}.
Recent advances in LLMs have demonstrated remarkable abilities in automating software development~\cite{gandhi2024budgetmlagent, tang2024codeagent} and code generation~\cite{guo2024dsagent, li2022competition, li2024autokaggle, nijkamp2023codegen}. Models like SWE-agent \citep{yang2024sweagent} and OpenAI’s Codex~\footnote{\url{https://openai.com/codex}} can translate natural language into program code with increasing proficiency. These successes, however, are mostly confined to textual programming languages, leaving a gap in support for graphical simulation and modeling environments~\cite{yang2024if, zheng2023survey}. In particular, existing LLM-based pipelines lack any native capability for Simulink model generation or manipulation. The fundamental difference between textual code and block-diagram models presents a unique challenge: LLMs trained predominantly on linear text struggle to represent and reason about the two-dimensional structure of diagrams \citep{abdalla2024generative}. Furthermore, unlike the wealth of open-source code available for training, there is a paucity of public data for Simulink or similar graphical models \citep{Shrestha2021SLGPT}. This data scarcity has hindered the specialization of LLMs for simulation tasks – early attempts like SLGPT (a GPT-2 fine-tuned on just 400 Simulink models) showed promise in generating simple block diagrams, but were limited by small corpus size \citep{Shrestha2021SLGPT} and could not generalize to complex systems. More recent efforts in the automotive domain fine-tuned 7B-parameter models for Simulink function generation and still had to synthesize training data via self-instruction due to the lack of real examples \citep{abdalla2024generative}. No prior work, to our knowledge, has achieved a general LLM-driven solution for automatically constructing full Simulink models from high-level specifications. Beginners usually replicate the now-existing Simulink diagram images to learn Simulink.
Inspired by this, we pose the following research question:
\textbf{Is it possible to automatically obtain a complete and correct simulation code from a textual simulation descriptions and the Simulink diagram images?}

To bridge the gap between advanced LLM capabilities and Simulink-based modeling, we introduce SimuGen, a novel multimodal agentic framework for automatic construction of Simulink simulation models. SimuGen leverages a state-of-the-art multimodal LLMs~\cite{openai2023gpt4, openai2023gpt4v} as its core reasoning engine, integrated with a suite of tools tailored for the Simulink environment. This agentic architecture ensures a modular, interpretable, and extensible workflow for universal simulation data generation. The contribution of this work can be summarized as:

\begin{enumerate}
    \item \textbf{Multimodal LLM Agent for Simulink Generation.} We propose the first framework that enables an LLM-based agent to interpret diagram images and generate corresponding Simulink models. Our system, \textit{SimuGen}, combines visual understanding of block diagrams with language-based reasoning and planning, emulating how human engineers translate diagrams into executable models.
    \item \textbf{Multi-Agent Workflow with Standard-Conformance Reviewer.} We design a multi-agent system incorporating a specialized unit test reviewer, inspired by SWE agents ~\citep{yang2024sweagent}. Unlike traditional unit tests that check code execution, our reviewer evaluates whether the investigator agent’s natural language-generated components align with Simulink modeling standards. 
    \item \textbf{Empirical Insights on Reviewer Effectiveness.} Case studies reveal that while the reviewer cannot directly assess the correctness of block connections, it can verify their compliance with modeling standards—thereby indirectly improving the overall accuracy and reliability of the generated models.
\end{enumerate}

\section{Challenges}
\label{sec:challenges}

Despite Simulink’s widespread use and its advantages in modeling physical systems through graphical interfaces, automating its usage with LLMs introduces significant challenges. Through our initial experiments, we observed that even advanced LLMs struggle with reliably generating valid Simulink models from either textual or visual inputs. These issues can be broadly categorized into three main types of errors, each pointing to a distinct deficiency in the model’s reasoning or implementation capabilities.

\subsection{Conceptual Errors: Missing Domain Knowledge}

These errors emerge when LLMs attempt to generate a Simulink model from a textual description without a sound understanding of domain-specific principles. For instance, an LLM may fail to grasp the mathematical structure underlying a control system or misinterpret the functional role of certain blocks in a feedback loop. Such \textbf{text-to-simulation} failures are often rooted in the model's lack of grounding in physical reasoning or engineering theory.

An illustrative example is provided in Appendix~\ref{sec:appendix-challenges}, where the LLM fails to properly measure the capacitor voltage in a Simscape-based RC circuit by directly tapping a conserving port rather than using a voltage sensor. This oversight reflects a misunderstanding of Simscape’s physical signal architecture.

To mitigate such issues, we propose using diagram inputs to help the LLM infer the structure of the model, allowing it to focus on retrieving domain-aligned block functions rather than generating topology from scratch. This approach explicitly embeds high-level intent and structural logic into the input, thereby improving conceptual coherence.

\subsection{Block Implementation Errors: Incorrect Block and Port Usage}

When generating MATLAB code to build models, LLMs often produce syntactically correct but semantically flawed outputs. These include referencing invalid block library paths, misconfiguring port numbers, or confusing block names—particularly for uncommon or custom components.

As illustrated in Figure~\ref{fig:agent-diagram}, these errors are commonly observed when users request LLMs to construct Simulink models via MATLAB code. Examples include incorrect block paths, mismatched ports, and duplicated connections, all of which contribute to broken model logic and execution failures.

\begin{figure*}[t]
    \centering
    \includegraphics[width=0.85\textwidth]{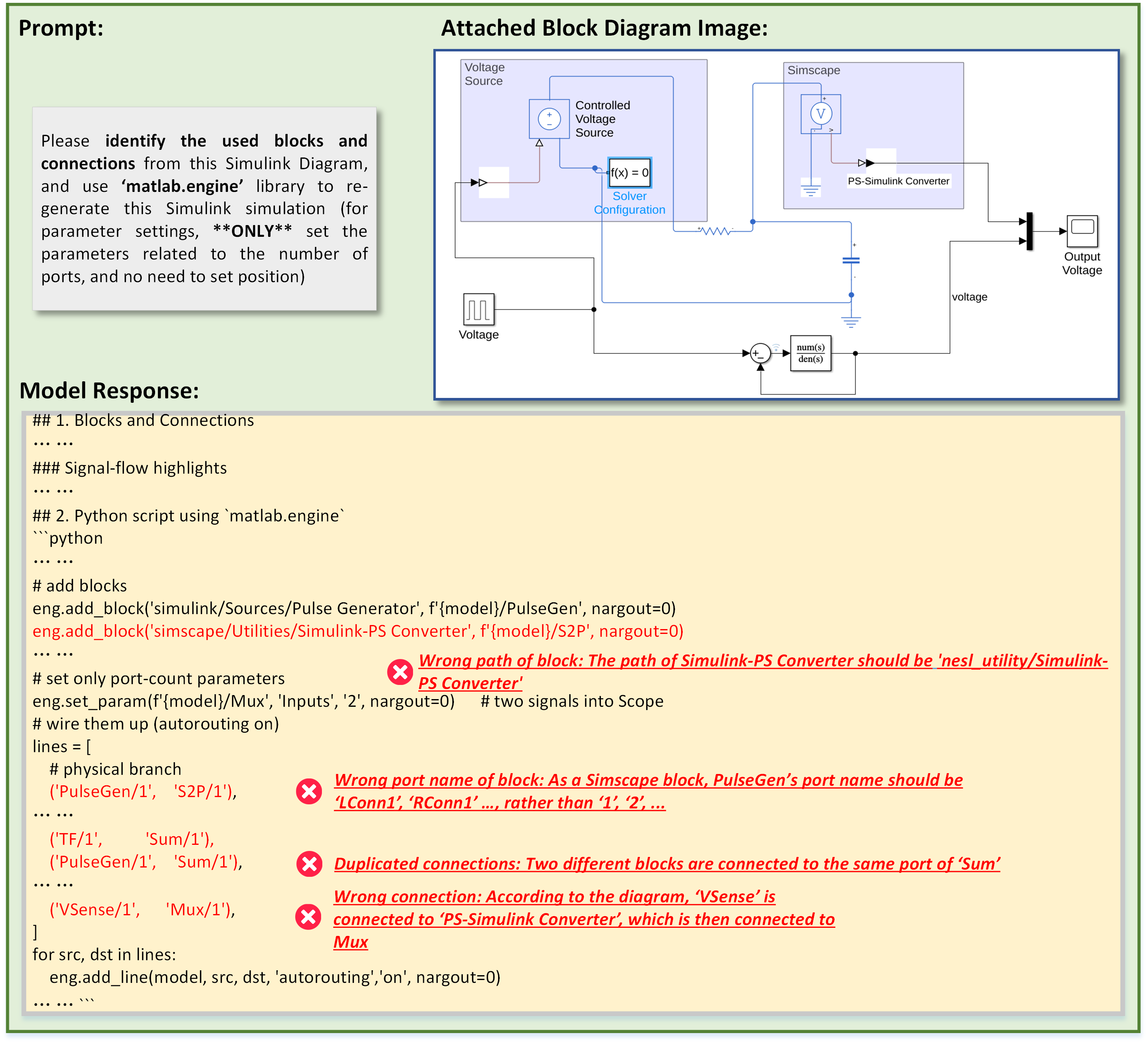}
    \caption{Common errors when users want LLM to generate MATLAB code for constructing Simulink model, such as incorrect block paths, port mismatches, and duplicated connections.}
    \label{fig:agent-diagram}
\end{figure*}

This category of errors reflects a lack of structured implementation knowledge. To mitigate this, we propose constructing a \textit{block description database} that provides LLMs with access to accurate, structured metadata about block functions, parameter names, port configurations, and usage examples.

\subsection{Linkage Implementation Errors: Invalid Connections}

In \textbf{diagram-to-simulation} workflows, LLMs frequently generate invalid wiring configurations that violate Simulink standards. Examples include connecting incompatible ports, omitting essential intermediary blocks, or mislabeling ports. These errors often bypass initial syntax checks but fail during execution, making them costly to debug.

To address this, we developed a \textit{Unit Test Reviewer} and \textit{Debug Locator} module to systematically evaluate connection integrity. The Unit Test Reviewer validates that all block connections adhere to Simulink specifications, while the Debug Locator assists in diagnosing faults by linking runtime errors to specific connection descriptions or block implementations.

\subsection{Context Window Bottlenecks in Single LLM Agents}

Recent studies have highlighted the limitations of single LLMs in handling long-context tasks. For instance, LLMs with even extended context windows still struggle with long input sequences, leading to performance degradation~\cite{li2024long}. Similarly, Fei et al. \cite{fei2024extending} proposed a semantic compression method to mitigate the challenges posed by long inputs, emphasizing the inherent difficulties faced by single LLMs. To address these issues, Zhang et al. \cite{zhang2024chain} introduced a multi-agent framework that enables better information aggregation and context reasoning across various LLMs.

Similarly, our investigation also revealed that using a single LLM to autonomously manage the entire Simulink automation pipeline—including block recognition, connection validation, code generation, and debugging—presents fundamental limitations. As the number of dialogue rounds increases and task types shift, the LLM’s contextual focus tends to degrade, leading to a decline in output quality. Furthermore, we observed a recurring pattern of overconfidence, where the LLM assumes the correctness of its generated connections without critically evaluating their validity. These issues result in persistent, hard-to-detect errors and expose the fragility of a single-agent workflow. All these factors motivate us to develop a multi-agent framework to solve the simulation automation tasks.

\section{Methodology}
\label{headings}
\subsection{SimuGen Framework Overview}

\begin{figure*}[t]
    \centering
    \includegraphics[width=\textwidth]{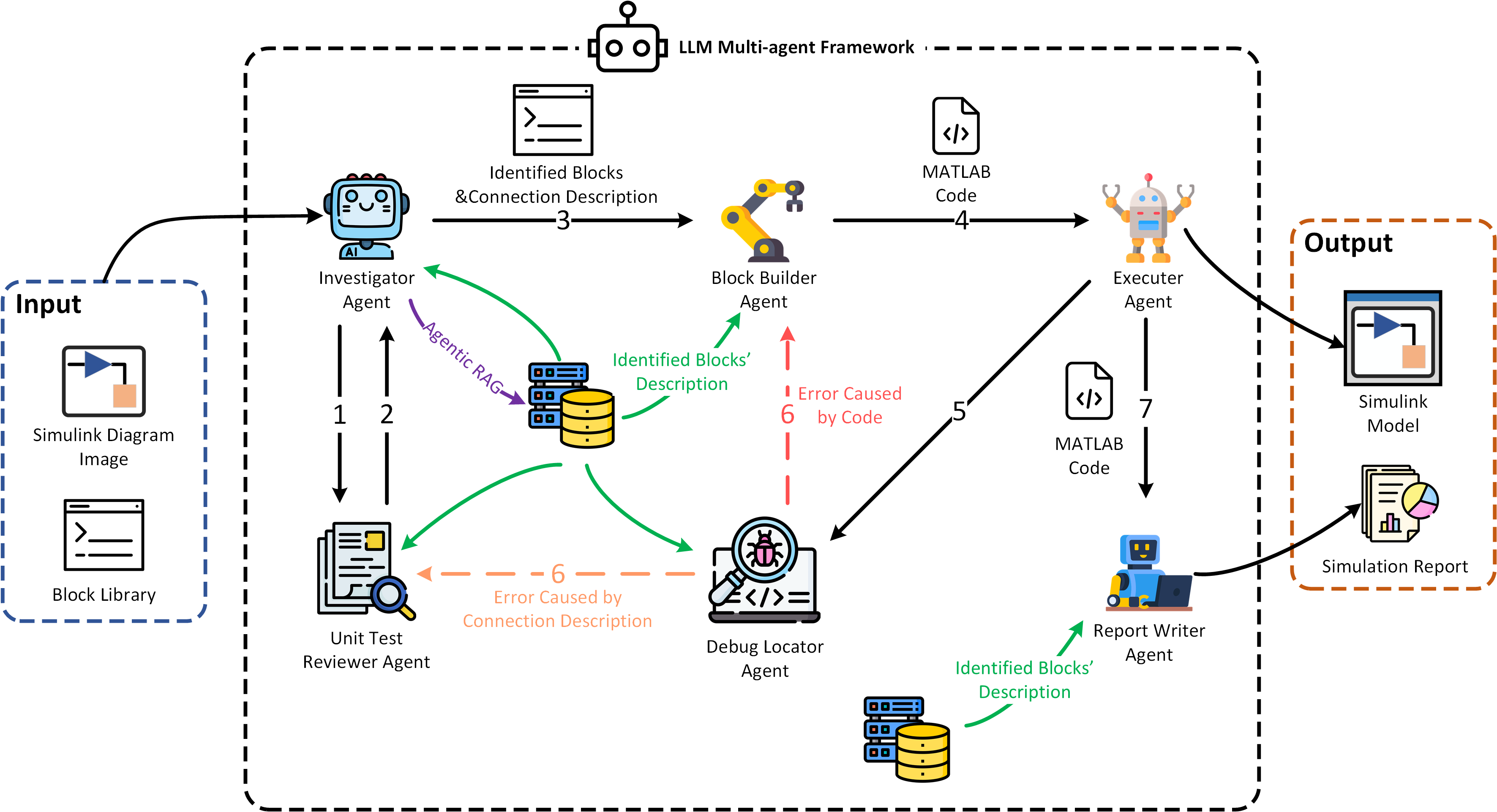}
    \caption{Overview of the agent-based architecture used in the SimuGen framework.}
    \label{fig:agent-diagram}
\end{figure*}

After identifying that current LLMs exhibit deficiencies in both conceptual understanding and implementation knowledge when generating Simulink simulations—as well as a notable decline in contextual focus as model complexity increases—we propose the SimuGen Framework to address these challenges, as shown in Figure~\ref{fig:agent-diagram}. Rather than relying on a single LLM to reason about the entire modeling process, which is prone to context window limitations and loss of focus, SimuGen employs a multi-agent system in which each agent is responsible for a specialized subtask. By decomposing the workflow, the multi-agent approach ensures that each agent maintains high task-specific attention, mitigating the risk of contextual degradation that plagues monolithic LLM solutions. Concretely, SimuGen takes a Simulink diagram image as input, providing explicit structural guidance and obviating the need for the LLM to infer model topology from scratch. The six specialized agents (Investigator, Unit Test Reviewer, Block Builder, Executor, Debug Locator, and Report Writer), together with a comprehensive Simulink block database, collaborate to bridge knowledge gaps and deliver robust, accurate simulation generation.

\subsection{Database Construction}

To provide agents with accurate and contextually relevant reference knowledge, we construct a comprehensive Simulink database that contains Full Block Descriptions, Function Descriptions, and Code Templates. The Full Block Description covers 50 commonly used Simulink blocks, detailing their block types, library paths, underlying principles, connection-related parameter specifications, and port names along with their descriptions. For a thorough explanation of the database construction process and illustrative examples, please refer to Appendix~\ref{sec:appendix-database}.

\subsection{LLM Agent Settings}
\paragraph{Investigator.}
The input to the Investigator consists of a Simulink diagram image and a block library containing 50 block types. The Investigator first identifies each block in the diagram image along with its corresponding block type. Leveraging an agentic retrieval-augmented generation (RAG) process, it then queries the database to retrieve the Full Block Descriptions for the relevant blocks. Subsequently, the Investigator outputs the connectivity relationships between blocks in the diagram image, formatted as: \textbf{BlockA (BlockA's block type) PortX <-> BlockB (BlockB's block type) PortY}.
 The complete Investigator prompt is provided in Appendix~\ref{sec:appendix-agent}.

\paragraph{Unit Test Reviewer.}
The Unit Test Reviewer receives as input the set of blocks and their connections identified by the Investigator within a given Simulink diagram image, along with the corresponding Full Block Descriptions. The Unit Test Reviewer assesses whether the proposed connections adhere to Simulink standards by performing eight checks:
1. \textbf{Identify the existence of block list};
2. \textbf{Identify any extra blocks};
3. \textbf{Formatting of block name};
4. \textbf{Formatting of connection description};
5. \textbf{Validate parameter settings in connections};
6. \textbf{Detect duplicate connections};
7. \textbf{Validate block connection types};
8. \textbf{Verify complete port connections}.
The complete Unit Test Reviewer prompt is provided in Appendix~\ref{sec:appendix-agent}.
   
\paragraph{Block Builder.}
 The Block Builder utilizes the Code Template, Function Description, and Full Block Description from the database to implement the simulation code for the diagram image in Simulink using matlab.engine, based on the connection descriptions provided by the Investigator. It is important to note that this work primarily focuses on the wiring aspects of simulations; therefore, the Block Builder is instructed to generate code exclusively with the add\_block and add\_line functions, and to use set\_param only when setting parameters related to the connections. The complete Block Builder prompt can be found in Appendix~\ref{sec:appendix-agent}.
 
\paragraph{Executor.}
 The Executor is an automated component responsible for running the code generated by the Block Builder. It executes the provided code and automatically returns the results of the execution.
 
\paragraph{Debug Locator.}
The Debug Locator analyzes error messages from the Executor to pinpoint the cause and identifies the relevant 5–10 lines of code for the Block Builder to modify. \textbf{In practice, we observed that most errors arise from logical flaws in the connection description that the Unit Test Reviewer fails to detect. Direct code modification in these cases would bypass the Investigator, resulting in code inconsistent with the original diagram}.

To resolve this, the Debug Locator first checks whether the code’s wiring and parameter settings match the connection description. If they match, the issue likely lies in the connection description itself, the Debug Locator then prompts the Unit Test Reviewer and Investigator to re-examine and revise it. If inconsistencies or syntax errors are found, feedback is sent directly to the Block Builder for code correction. The complete Debug Locator prompt is provided in Appendix~\ref{sec:appendix-agent}.

\paragraph{Report Writer.}
After the simulation has executed successfully, the Report Writer generates a comprehensive simulation report by synthesizing Full Block Descriptions, connection descriptions, and code implementation details. Specifically, the report generated by the Report Writer is structured into four sections: \textbf{(1) What is the simulation about? (2) What are the main simulation steps? (3) What theoretical knowledge and mathematical modeling are involved in each step? (4) How is each step implemented in code?} The complete Report Writer prompt can be found in Appendix~\ref{sec:appendix-agent}.

\subsection{Full Agentic Workflow}
The overall workflow proceeds as follows:

1. The Investigator receives a Simulink diagram and a block library, identifies present blocks, retrieves their Full Block Descriptions from the RAG database, and extracts connection descriptions.

2. The Unit Test Reviewer checks whether the connection descriptions comply with the Simulink rules. If violations are found, the Investigator revises the connection descriptions based on the reviewer's feedback.

3. If valid, the Investigator passes the descriptions to the Block Builder, who generates simulation code using the matlab.engine library.

4. The executor will execute the code passed from the Block Builder.

5. For execution errors, the Debug Locator determines if the issue is due to code generation or missed logical errors, providing targeted feedback to the Block Builder or Investigator for further revision.

6. After successful execution, the Report Writer compiles the latest block descriptions, connection details, and code into a comprehensive simulation report.

\section{Experiments}
\label{others}
\subsection{Experimental Setup}

\begin{table*}[t]
  \centering
  \caption{Simulation domains and complexity of ground truth (GT) models for 9 simulation examples.}
  \setlength{\tabcolsep}{4pt} 
  \begin{tabular}{l l cc}
    \toprule
    \textbf{Simulation Domain}                    & \textbf{Simulation Name}                   & \textbf{GT Blocks} & \textbf{GT Connections} \\
    \midrule
    \multirow{4}{*}{General Application} & Artificial Algebraic Loops   & 5          & 5               \\
                                         & State Event                  & 5          & 6               \\
                                         & Zero Crossing Detection      & 9          & 10              \\
                                         & Simulink Model               & 11         & 13              \\
    \midrule
    Physical Modeling   & Bouncing Ball                & 7          & 7               \\
    \midrule
    \multirow{3}{*}{Electrical Systems Modeling} & Bipolar Transistor     & 13         & 20              \\
                                         & RC Circuit                   & 14         & 16              \\
                                         & 2 Bus Loadflow               & 14         & 13              \\
    \midrule
    Automotive Modeling                  & Wheel Speed                  & 12         & 11              \\
    \bottomrule
  \end{tabular}
  \label{table:accuracy}
\end{table*}

\paragraph{Task Selections.}
To comprehensively evaluate the performance of SimuGen, we selected nine Simulink simulations spanning four representative domains: General Applications (covering artificial algebraic loops, Simulink models, state events, and zero-crossing detection), Physical Modeling (bouncing ball), Electrical System Modeling (including 2-bus load flow, bipolar transistor, and RC circuit), and Automotive Modeling (wheel speed).
Table 1 summarizes the complexity of the ground truth (GT) models corresponding to these simulation examples. In our study, model complexity is quantified by the number of blocks and connections present in each GT model.

\paragraph{LLM Selections.}
Our framework involves five agents in total. Among them, the Investigator Agent is responsible for both visual recognition and visual reasoning tasks, while the remaining agents are focused solely on reasoning. Accordingly, we selected GPT-4.1 and o4-mini as the LLM backbones for the Investigator Agent.
GPT-4.1 was chosen due to its status as OpenAI's flagship model and its state-of-the-art performance in multimodal benchmarks. o4-mini, on the other hand, was selected for its support of vision-integrated chain-of-thought reasoning, a feature we hypothesize to be particularly beneficial for the Investigator Agent’s interaction with the Unit Test Reviewer Agent.
For the other agents, all of which primarily require reasoning capabilities, we use o3-mini as the underlying LLM, striking a balance between performance and computational efficiency.

\paragraph{Accuracy of the Generation.} 
To evaluate the accuracy of the generated simulations produced by the framework, we define the overall accuracy as the average of the block accuracy and connection accuracy, calculated as follows:

\begin{equation}
\text{Accuracy} = \frac{1}{2} \left( \frac{|B_{\text{match}}|}{|B_{\text{GT}}|} + \frac{|C_{\text{match}}|}{|C_{\text{GT}}|} \right)
\end{equation}

where:
\begin{itemize}
    \item $B_{\text{GT}}$ is the set of blocks in the ground truth (GT) model.
    \item $B_{\text{gen}}$ is the set of blocks in the generated model.
    \item $B_{\text{match}} = B_{\text{GT}} \cap B_{\text{gen}}$ is the set of correctly predicted blocks, i.e., blocks that exist in both the GT and generated models.
    \item $C_{\text{GT}}$ is the set of connections (e.g., edges or links) in the ground truth model.
    \item $C_{\text{gen}}$ is the set of connections in the generated model.
    \item $C_{\text{match}} = C_{\text{GT}} \cap C_{\text{gen}}$ is the set of correctly predicted connections, i.e., connections that exist in both the GT and generated models.
\end{itemize}

\begin{table}[t]
  \centering
  \caption{Accuracy (\%) on 9 simulation examples: Investigator agent using o4-mini/GPT-4.1, others use o3-mini.}
  \begin{tabular}{@{} llc @{}}
    \hline
    \textbf{Method}                & \textbf{o4-mini} & \textbf{GPT-4.1} \\
    \hline
    Artificial Algebraic Loops     & 100              & 100              \\
    State Event                    & 73.3             & 68.55            \\
    Zero Crossing Detection        & 95.5             & 72.25            \\
    Simulink Model                 & 100              & 87.75            \\
    Bouncing Ball                  & 100              & 100              \\
    Bipolar Transistor             & 95               & 77.5             \\
    RC Circuit                     & 95.3             & 71.88            \\
    2 Bus Loadflow                 & 100              & 88.7             \\
    Wheel Speed                    & 91.25            & 82.15            \\
    \hline
    \textbf{Average}               & \textbf{94.5}    & \textbf{83.2}    \\
    \hline
  \end{tabular}
  \label{tab:accuracy_single}
\end{table}

\begin{table}[t]
  \centering
  \caption{Cost and run time for 9 simulation examples: Investigator agent uses o4-mini, others use o3-mini.}
  \begin{tabular}{@{} llc @{}}
    \hline
    \textbf{Simulation Example}    & \textbf{Cost (\$)} & \textbf{Run Time (s)} \\
    \hline
    Artificial Algebraic Loops     & 0.06              & 121.91              \\
    State Event                    & 0.05              & 105.99              \\
    Zero Crossing Detection        & 0.09              & 169.57              \\
    Simulink Model                 & 0.13              & 232.35              \\
    Bouncing Ball                  & 0.22              & 297.34              \\
    Bipolar Transistor             & 0.41              & 642.44              \\
    RC Circuit                     & 0.16              & 358.27              \\
    2 Bus Loadflow                 & 0.14              & 221.89              \\
    Wheel Speed                    & 0.21              & 325.95              \\
    \hline
    \textbf{Mean}                  & \textbf{0.17}     & \textbf{275.41}    \\
    \hline
  \end{tabular}
  \label{tab:cost_runtime_single}
\end{table}

\subsection{Main Results}
From Table~\ref{tab:accuracy_single}, we observe that the Investigator agent successfully completes all 9 tasks when using either o4-mini or GPT-4.1. However, when powered by o4-mini, the agent achieves a Simulink simulation reproduction accuracy of $94.5\%$, compared to $83.2\%$ with GPT-4.1. We hypothesize that this performance gap arises from o4-mini's enhanced visual reasoning capabilities. Specifically, the Investigator agent equipped with o4-mini tends to zoom in on the image regions highlighted by the Unit Test Reviewer in their feedback, leading to more accurate identification of the corresponding blocks and connections in the simulation diagram.

We further observe that regardless of which model the Investigator agent uses, the reproduction accuracy tends to be inversely correlated with the complexity of the ground truth (GT) Simulink model. In cases where the number of connections is fewer than 10, the reproduced models generally achieve an accuracy close to $100\%$. An exception to this trend is the \textit{State Event} case, where the accuracy drops significantly. We hypothesize that this may be due to limited exposure of both o4-mini and GPT-4.1 to high-quality images of various Simulink blocks during pretraining, particularly those related to event-based dynamics.

In terms of costing, Table~\ref{tab:cost_runtime_single} summarizes the monetary cost (in USD) and execution time (in seconds) for nine simulation examples processed by our framework. In each case, the Investigator agent utilizes the o4-mini model, whereas the remaining agents use o3-mini. Results demonstrate that typical simulation tasks can be completed within approximately \$0.17 and 275 seconds on average.

\begin{table*}[t]
  \centering
  \caption{Performance (\%) of SimuGen and its ablations on 9 simulation examples: Investigator agent using o4-mini and all other agents using o3-mini}
  \setlength{\tabcolsep}{4pt} 
  \begin{tabular}{lrrrr}
    \toprule
    \textbf{Method}                              & \textbf{SimuGen} & \textbf{w/o Unit Test Reviewer} & \textbf{w/o Debug Locator} & \textbf{w/o All} \\
    \midrule
    Artificial Algebraic Loops          & 100     & 100                     & 100               & 100     \\
    State Event                         & 73.3    & 73.3                    & 73.3              & 73.3    \\
    Zero Crossing Detection             & 95.5    & 95.5                    & 95.5              & 95.5    \\
    Simulink Model                      & 100     & 100                     & 100               & 100     \\
    Bouncing Ball                       & 100     & 92.8                    & 100               & --      \\
    Bipolar Transistor                  & 95      & 61.65                   & --                & --      \\
    RC Circuit                          & 95.3    & 95.3                    & 95.3              & --      \\
    2 Bus Loadflow                      & 100     & 96.15                   & 100               & 96.15   \\
    Wheel Speed                         & 91.25   & 91.25                   & 91.25             & --      \\
    \midrule
    \textbf{Average}                    & \textbf{94.5}    & \textbf{86.21}            & \textbf{83.9}     & \textbf{51.7} \\
    \bottomrule
  \end{tabular}
  \label{table:simugen_ablations}
\end{table*}

\subsection{Alblation Study}

‘In our ablation study, we evaluate three variants of the system: (1) without the Unit Test Reviewer, (2) without the Debug Locator, and (3) without both components. It is important to note that in the absence of the Debug Locator, the code is executed only once, and any runtime error is treated as a task failure. Overall, from Table 3, we observe that excluding either the Unit Test Reviewer or the Debug Locator leads to a noticeable drop in simulation reproduction accuracy. Notably, when both components are removed, the accuracy drops significantly to $51.7\%$. This highlights a synergistic effect between the Unit Test Reviewer and the Debug Locator, indicating that their combined contribution to SimuGen is greater than the sum of their individual parts. MEanwhile, we observe that for the \textit{Bouncing Ball}, \textit{RC Circuit}, and \textit{Wheel Speed} simulations, none can be completed in the absence of both the Unit Test Reviewer and the Debug Locator agents. This result highlights that, although the Unit Test Reviewer and Debug Locator operate via distinct mechanisms to reflect and reproduce the Simulink modeling process, both play an essential role in ensuring successful reconstruction of Simulink models.

It is also worth noting that in the \textit{Bipolar Transistor} case, SimuGen achieves an accuracy that is $33.35\%$ higher than the variant without the Unit Test Reviewer. This suggests that although the Unit Test Reviewer cannot directly assess whether the Investigator's predicted connections are correct, it can still provide valuable feedback on the logical coherence of the connections. Such feedback encourages the Investigator to re-examine the identified blocks and connections, ultimately leading to more accurate reproductions.

At the same time, we observe that for most of the 9 tasks, simulations can still be successfully reproduced even without the Debug Locator. This indicates that as long as the blocks and connections identified by the Investigator are logically sound, generating the corresponding executable code is relatively straightforward. However, in the \textit{Bipolar Transistor} case, simulation fails when the Debug Locator is removed. This underscores the complementary role of the Debug Locator: in scenarios where the Unit Test Agent overlooks potential issues, the Debug Locator can interact with the execution environment and analyze runtime outputs to compensate for such oversights.

\section{Related works}

\paragraph{LLMs for Code Generation and Mathematical Reasoning.}~Large language models trained on code have demonstrated impressive capabilities in both program synthesis and multi‐step reasoning.
Chen et al. \cite{chen2021evaluating} evaluated code‐trained LLMs on a diverse set of programming tasks, showing they can generate nontrivial functions with high accuracy. 
Li et al. \cite{li2022competition} introduced AlphaCode, which achieved human‐comparable performance on competitive programming benchmarks by combining a powerful generative model with search techniques. 
Chain‐of‐thought prompting has further improved LLM's reasoning by eliciting intermediate logical steps, leading to better results on quantitative problems~\cite{wei2022chain}. The collaboration of multiple LLMs was proposed for complex data science applications~\cite{hong2024data, jing2024dsbench, li2024autokaggle, jimenez2023swebench}.

% \paragraph{Challenges in Simulation Code Generation.}~Despite their success in abstract code tasks, LLMs often falter when asked to generate domain-specific simulation scripts from scratch. 
% \citet{meadows2024exploring} demonstrated that LLMs’ physics inference breaks down when key premises are missing, leading to incorrect or incomplete models. 
% \citet{gu2025effectiveness} analyzed LLMs on domain-specific code generation and found frequent semantic errors in simulation-related scripts without targeted training data.

\paragraph{Data Scarcity in Simulation Domains.}~ Liu et al. \cite{liu2024toward, liu2024towards} proposed LLMs to generate Python codes for simulation tasks.
However, many fields rely on the Simulink platform for their simulations.
Luitel et al. \cite{Luitel2024simulink} applied LLMs to slice and analyze Simulink models. 
However, they did not propose a comprehensive system for generating and conducting simulations.

\paragraph{Multimodal Agentic Frameworks.}  

Li et al. \cite{li2023camel} introduced the first agent systems based on LLMs that facilitate cooperative interactions among multiple agents. 
In contrast, another form of agent interaction is adversarial~\citep{gou2023critic}. 
To address multimodal tasks, a hierarchical multimodal retrieval-augmented generation framework was designed for multi-agent systems focused on perception, reasoning, and generation \cite{liu2025hm}. 
Qian et al. \cite{qian2024chatdev} developed ChatDev, wherein communicative agents collaboratively drive the software development lifecycle, illustrating the benefits of modular agentic workflows. 

Our SimuGen framework builds on these ideas by assigning distinct roles—Investigator, Unit Test Reviewer, Block Builder, Executor, Debug Locator, and Report Writer—to reliably generate and validate simulation code.

\section{Conclusion}

SimuGen advances the automated generation of Simulink models by introducing a collaborative multi-agent framework that leverages both visual and domain knowledge. Through specialized agents for diagram interpretation, validation, code generation, and debugging—including a standard-conformance unit test reviewer that assesses whether model components adhere to Simulink modeling standards—SimuGen reliably transforms Simulink diagram images into accurate, executable simulation code. Experimental results show that SimuGen achieves an average reproduction accuracy of 94.5\% across diverse simulation tasks. Case studies further indicate that, while the reviewer cannot directly assess the correctness of block connections, its ability to enforce modeling standards indirectly improves the overall reliability of the generated models. These results demonstrate the effectiveness of a modular, agent-based approach with built-in standard-conformance checks for Simulink model construction.

\bibliographystyle{plain}
\bibliography{reference}

\appendix

\clearpage

\counterwithin{table}{section}
\counterwithin{figure}{section}
\counterwithin{equation}{section}

\lstset{
  basicstyle=\ttfamily\footnotesize,
  columns=fullflexible,
  breaklines=true,
  frame=single,
  backgroundcolor=\color{gray!10},
  keywordstyle=\color{blue},
  commentstyle=\color{green!60!black},
  stringstyle=\color{black},         % <--- safer than orange
  rulecolor=\color{black},           % <--- sets frame color
  showstringspaces=false,
  captionpos=b
}

\section{Appendix: Challenges}
\label{sec:appendix-challenges}
\renewcommand{\thesection}{\Alph{section}}
To illustrate common conceptual errors made by LLMs when generating Simulink models, we present an example based on a user's request to build a simple RC circuit using Simscape blocks. The generated code successfully constructs the topology using \texttt{matlab.engine}, but fails to include a necessary voltage sensor for signal measurement.

\subsection*{User Query}
\begin{itshape}
Could you generate a simulation \emph{RC circuit} using Simscape library from Simulink? Please follow the following steps:

1. Provide the blocks needed and their connections (there should be a Scope to observe the result).

2. Generate the simulation code using \texttt{matlab.engine}.

\textbf{Note:} Only generate blocks and connections; there is no need to set parameters.
\end{itshape}

\subsection*{Blocks and Library Paths}

\begin{table}[h]
\centering
\caption{Blocks used in the generated RC circuit model}
\small
\begin{tabularx}{\linewidth}{lX}
\toprule
\textbf{Block Name} & \textbf{Library Path} \\
\midrule
Solver Configuration & Simscape > Commonly Used Blocks \\
Electrical Reference & Simscape > Foundation Library > Electrical Elements \\
DC Voltage Source & Simscape > Electrical Sources > DC Voltage Source \\
Resistor & Simscape > Electrical Elements > Resistor \\
Capacitor & Simscape > Electrical Elements > Capacitor \\
PS-Simulink Converter & Simscape > Utilities > PS-Simulink Converter \\
Scope & Simulink > Sinks > Scope \\
\bottomrule
\end{tabularx}

\label{tab:rc-blocks}
\end{table}

\subsection*{Connection Overview}

\noindent The following is a textual representation of the circuit structure:

\medskip

\begin{tabular}{l}
\texttt{Vsrc/1  -->  R/1} \\
\texttt{R/2     -->  C/1} \\
\texttt{C/2     -->  Ground/1} \\
\texttt{Ground/1 --> SolverConfig/1} \\
\texttt{Vsrc/2 --> Ground/1} \\
\texttt{C/1     --> PS2SL/1} \\
\texttt{PS2SL/1 --> Scope/1}
\end{tabular}

\medskip

This design routes the voltage across the capacitor directly into the Scope via the PS-Simulink Converter.

\clearpage
\subsection*{Generated Code Using \texttt{matlab.engine}}
\begin{lstlisting}[language=Python, caption={LLM-generated Python code to build the RC circuit model}]
import matlab.engine

# Start MATLAB engine
eng = matlab.engine.start_matlab()

# Create and open model
model = 'rc_circuit'
eng.new_system(model, 'Model')
eng.open_system(model)

# Add blocks
blocks = {
    'SolverConfig': 'simscape/Commonly Used Blocks/Solver Configuration',
    'Ground':       'simscape/Foundation Library/Electrical/Electrical Elements/Electrical Reference',
    'Vsrc':         'simscape/Foundation Library/Electrical/Electrical Sources/DC Voltage Source',
    'R':            'simscape/Foundation Library/Electrical/Electrical Elements/Resistor',
    'C':            'simscape/Foundation Library/Electrical/Electrical Elements/Capacitor',
    'PS2SL':        'simscape/Utilities/PS-Simulink Converter',
    'Scope':        'simulink/Sinks/Scope'
}

for name, path in blocks.items():
    eng.add_block(path, f'{model}/{name}')

# Make connections
eng.add_line(model, 'Vsrc/1',     'R/1',        'autorouting','on')
eng.add_line(model, 'R/2',        'C/1',        'autorouting','on')
eng.add_line(model, 'C/2',        'Ground/1',   'autorouting','on')
eng.add_line(model, 'Ground/1',   'SolverConfig/1','autorouting','on')
eng.add_line(model, 'Vsrc/2',     'Ground/1',   'autorouting','on')
eng.add_line(model, 'C/1',        'PS2SL/1',    'autorouting','on')
eng.add_line(model, 'PS2SL/1',    'Scope/1',    'autorouting','on')

# Save model
eng.save_system(model)
# eng.sim(model)  # Uncomment to simulate
\end{lstlisting}

\subsection*{Conceptual Issue Identified}

Although the generated structure is syntactically correct, it fails conceptually: Simscape does not allow direct measurement from conserving electrical ports. Instead, a \textbf{Voltage Sensor} block must be used to measure across the capacitor, with its output routed to a PS-Simulink Converter and then to the Scope.

\textbf{Summary:} The LLM omits the essential measurement interface, demonstrating a conceptual gap in how physical signal flows work in Simscape. This is a typical example of domain-specific misunderstanding, even when code generation succeeds syntactically.

\clearpage
\section{Appendix: Datasets}
\label{appendix:datasets}

\begin{figure*}[h]
    \centering
    \includegraphics[width=0.85\textwidth]{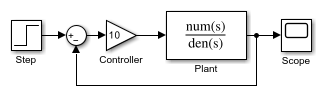}
    \caption{Case study: artificial algebraic loops.}
\end{figure*}

\begin{figure*}[h]
    \centering
    \includegraphics[width=0.85\textwidth]{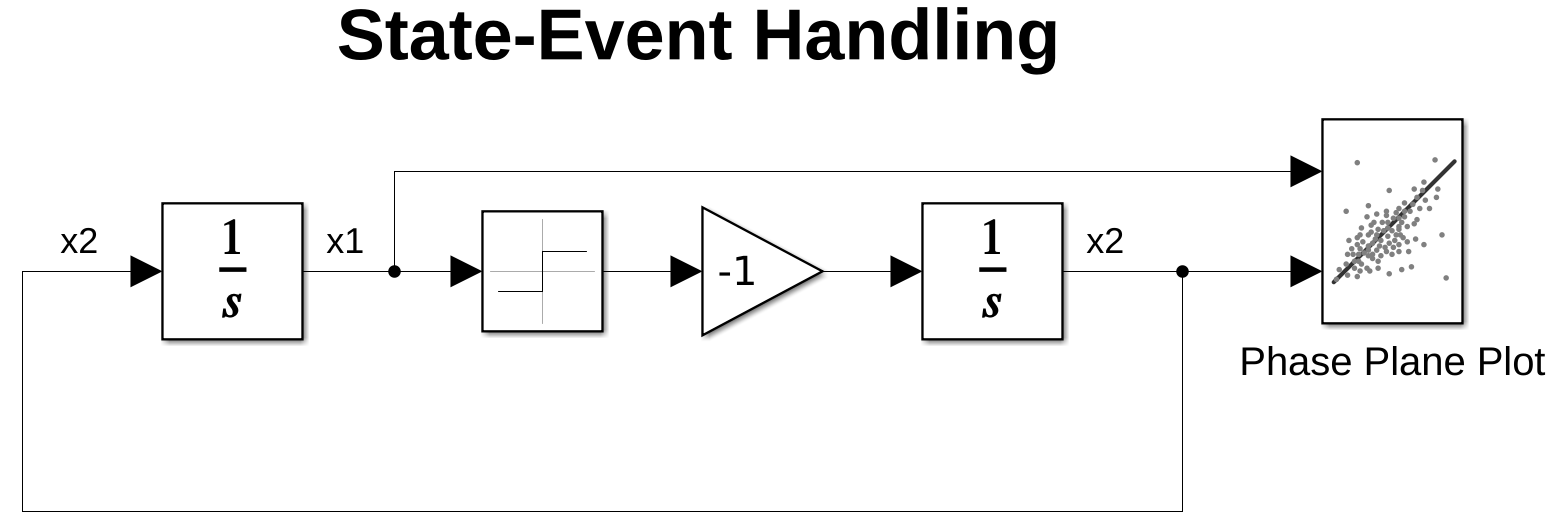}
    \caption{Case study: state event.}
\end{figure*}

\begin{figure*}[h]
    \centering
    \includegraphics[width=0.85\textwidth]{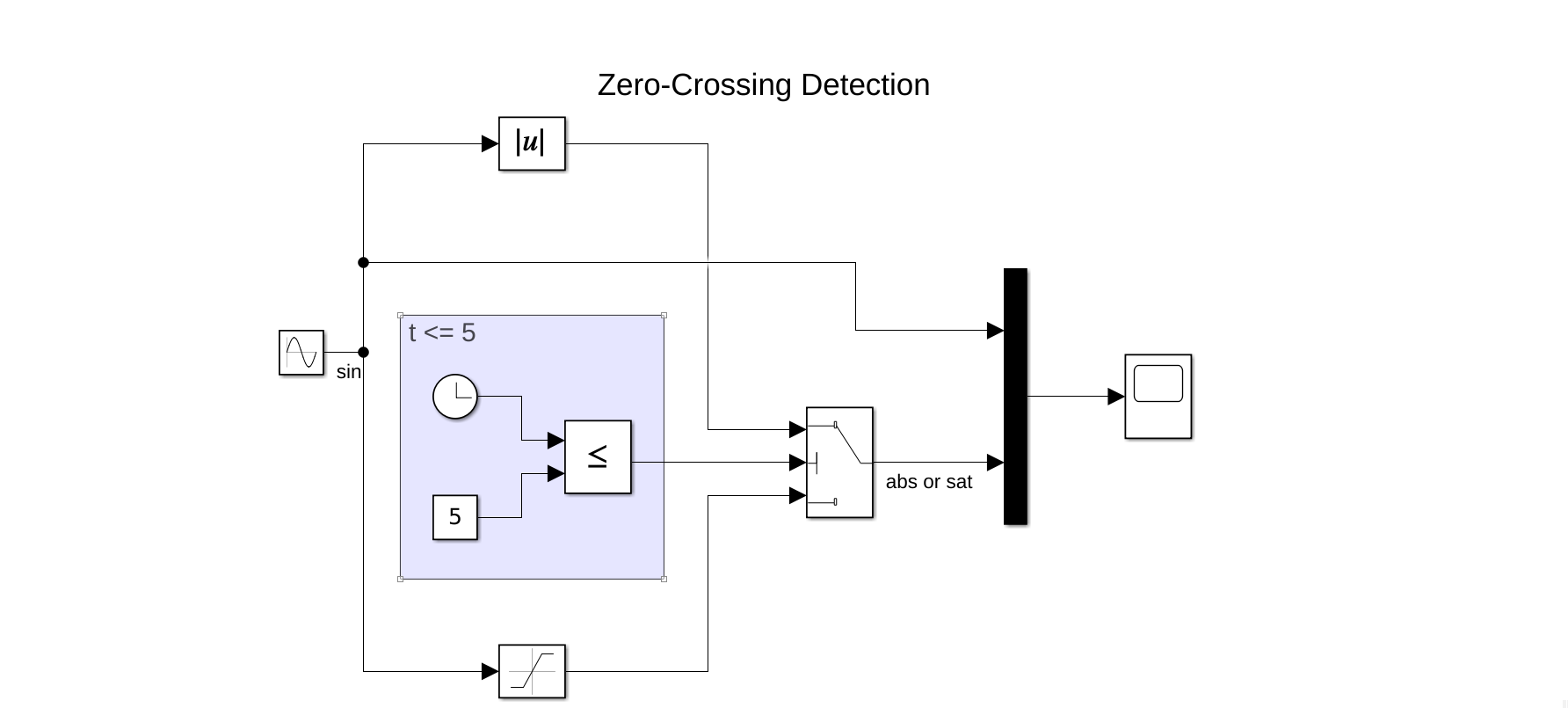}
    \caption{Case study: zero crossing detection.}
\end{figure*}

\begin{figure*}[h]
    \centering
    \includegraphics[width=0.85\textwidth]{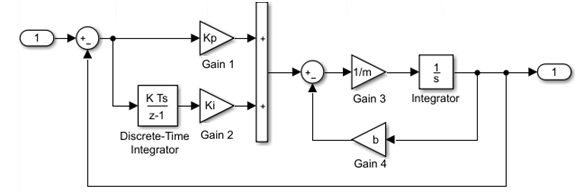}
    \caption{Case study: simulink model.}
\end{figure*}

\begin{figure*}[h]
    \centering
    \includegraphics[width=0.85\textwidth]{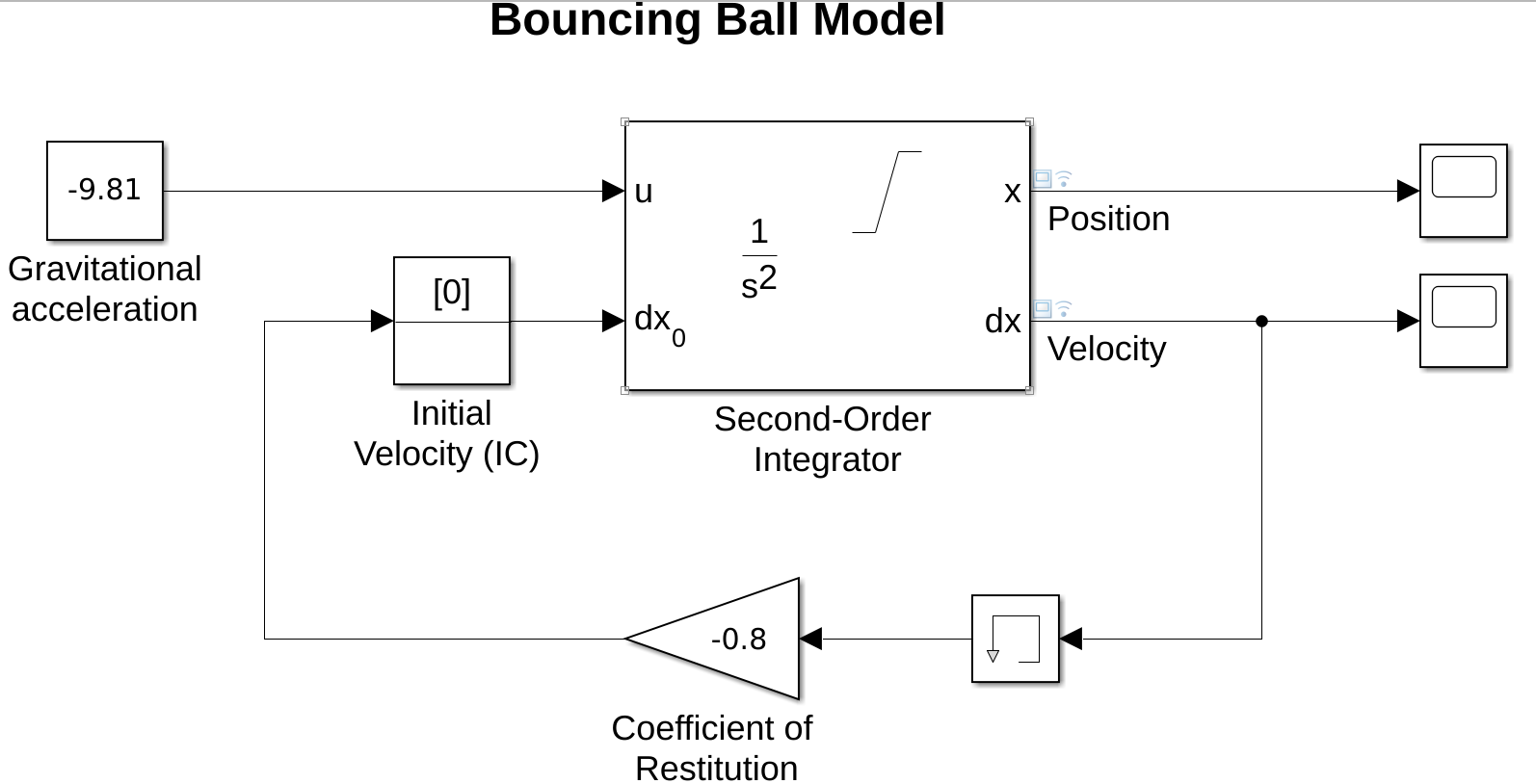}
    \caption{Case study: bouncing ball.}
\end{figure*}

\begin{figure*}[h]
    \centering
    \includegraphics[width=0.85\textwidth]{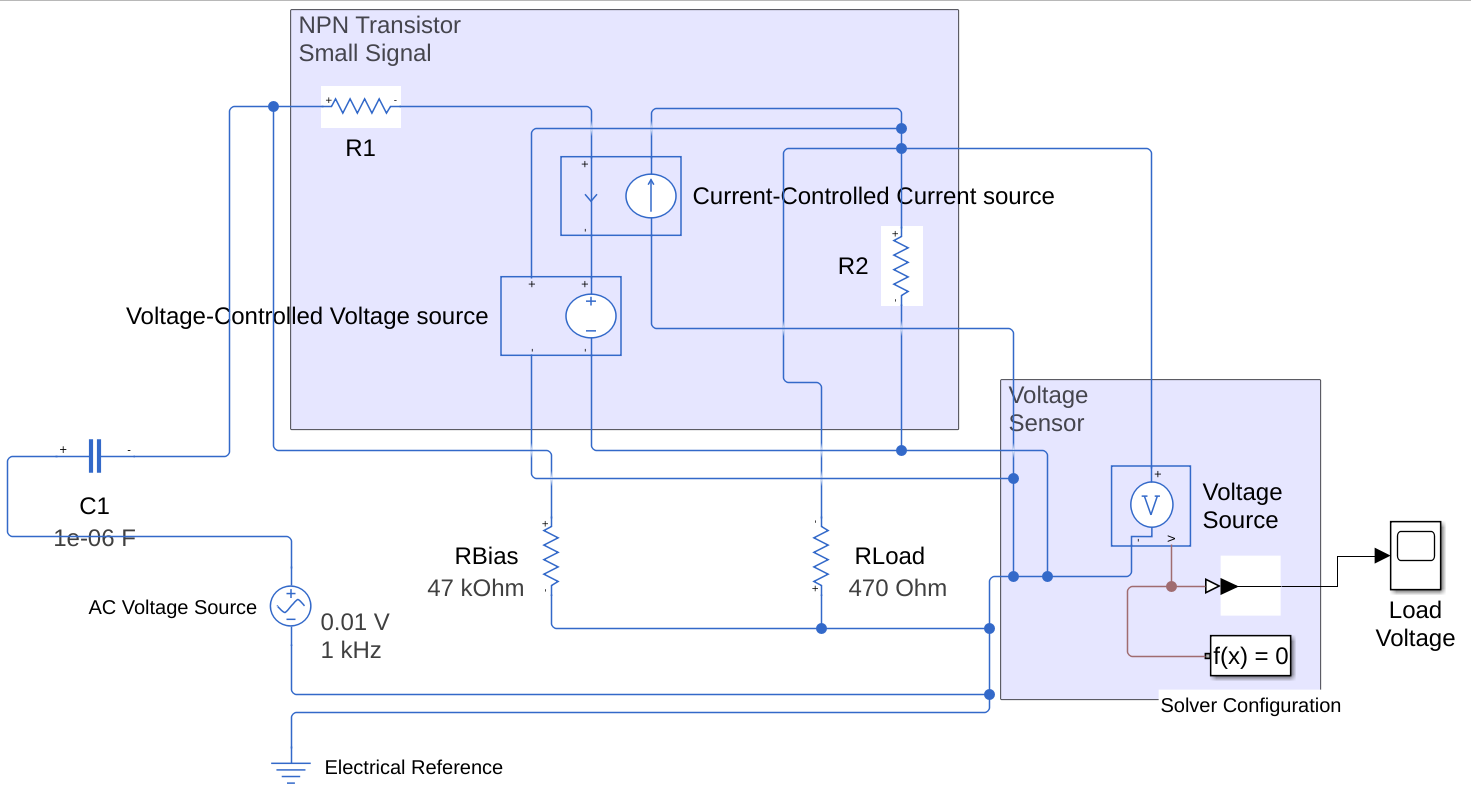}
    \caption{Case study: bipolar transistor.}
\end{figure*}

\begin{figure*}[h]
    \centering
    \includegraphics[width=0.85\textwidth]{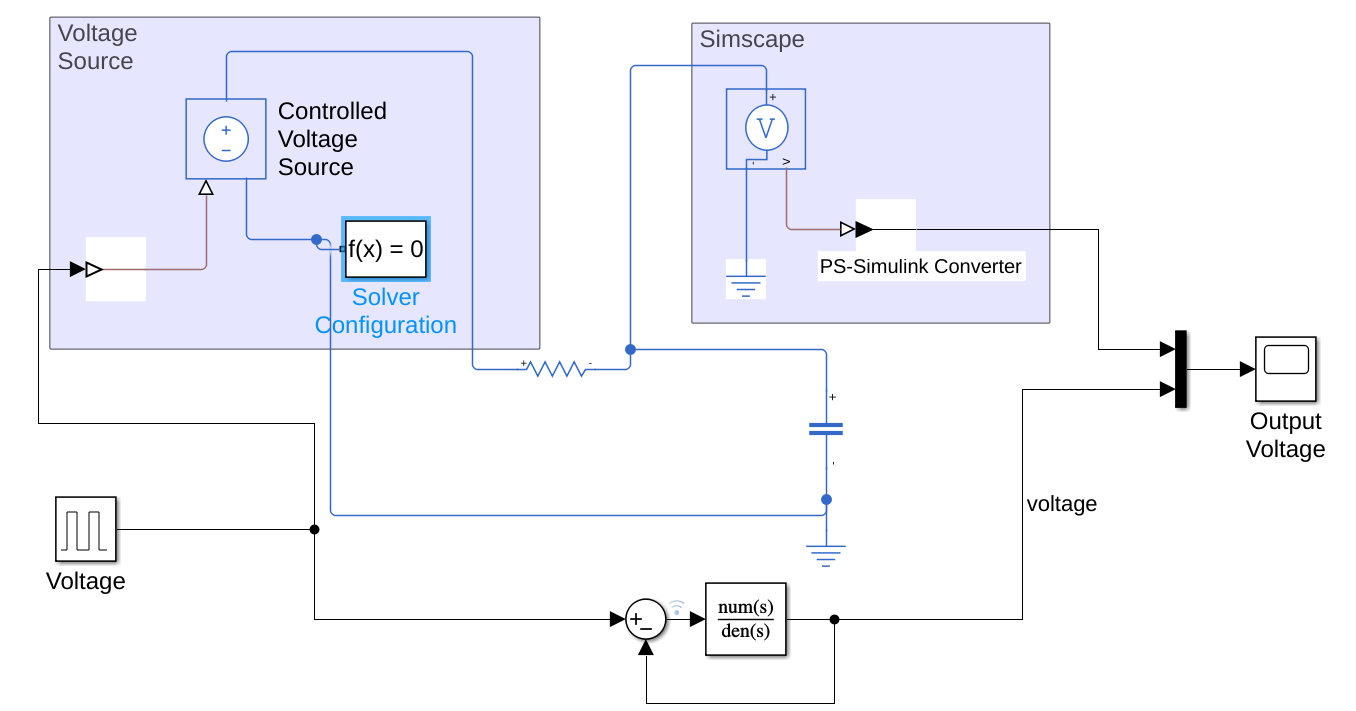}
    \caption{Case study: RC circuit.}
\end{figure*}

\begin{figure*}[h]
    \centering
    \includegraphics[width=0.85\textwidth]{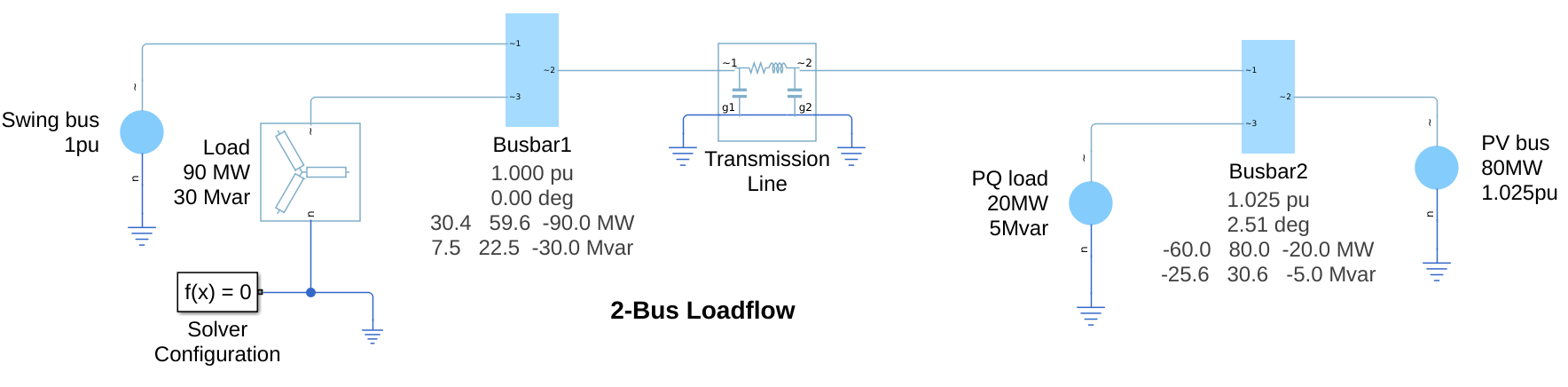}
    \caption{Case study: two bus.}
\end{figure*}

\begin{figure*}[h]
    \centering
    \includegraphics[width=0.85\textwidth]{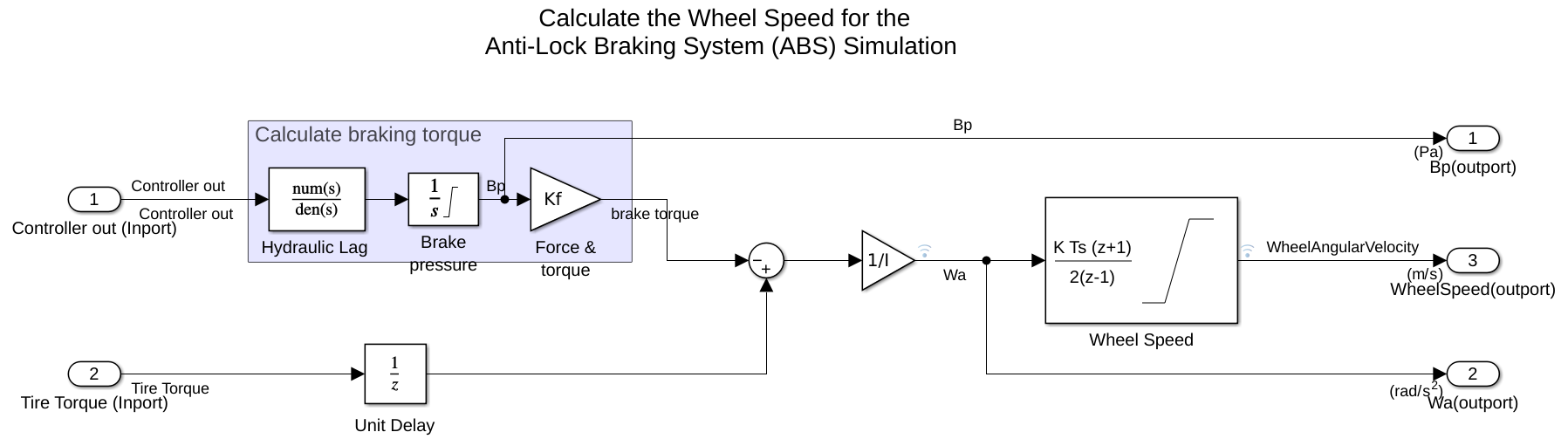}
    \caption{Case study: wheel speed.}
\end{figure*}

\clearpage
\section{Appendix: Prompts of Agents}
\label{sec:appendix-agent}
\begin{lstlisting}[language=bash, caption={Prompt of the Investigator agent in round 1.}]
# ROLE #
You are an Investigator responsible for deeply analyzing a simulation explanation, its corresponding simulation diagram, and a simulation blocks list.

# TASK #
You will receive a detailed simulation explanation, its corresponding simulation diagram, and a simulation blocks list that includes the block types. Your tasks are to:
- List all the blocks used in the simulation, **INCLUDING those without annotations if they are USED**. Before listing them, identify their corresponding block types from the simulation blocks list, **while please note that the diagram would NEVER include a SUBSYSTEM, so DO NOT request a subsystem.** The list should be formatted as:

BlockA (BlockA's block type)  
BlockB (BlockB's block type)  
...

where Used BlockA, BlockB are just example, if the annotation of a block is already given, please just use it. **BUT NEVER include the special symbol '/' in the block name, if it happen in the annotation, you need to modified it.**

- **Important:** If the simulation diagram shows multiple instances of the same block without separate annotations, you should proactively annotate them to differentiate each instance. For example, there are two sum without annotation are shown in the diagram, you should list them like:

sum 1 (sum)
sum 2 (sum)


# INPUT #
- **Simulation Explanation:**  
  {simulation_explanation}

- **Simulation Blocks List:**  
  {simulation_blocks_list}
  
# RESPONSE #
Always list all identified blocks first and then request half description of **ALL** listed blocks using their **block type (not Path) from the blocks list** at the end of your response in a JSON format as (DO NOT need to request the same unique block type):

{
  "request_blocks": ["block type 1", "block type 2", ...]
}

Do not ask any clarifying questions or confirmations. Directly provide a complete answer.
\end{lstlisting}

\clearpage
\begin{lstlisting}[language=bash, caption={Prompt of the Investigator agent in round 2.}, basicstyle=\fontsize{9pt}{9pt}\ttfamily\selectfont]
# INPUT #

- **Simulation Blocks Description:**  
  {simulation_blocks_list}

# TASK #
According to the **block you identified in the simulation diagram**, you will receive an identified block's **Simulation Blocks Description** that includes their block type and port description. Your tasks are to:

- Clearly describe the connections and data flow among these blocks in the simulation diagram using their block types. First, determine which port labels are connected, then match them to the corresponding real port names from the **Simulation Blocks Description**. **MUST STRICTLY Format** each connection description as:  

  BlockA (BlockA's block type) PortX (**related parameter setting to match the port number if necessary**)<-> BlockB (BlockB's block type) PortY (**related parameter setting to match the port number if necessary**)
  
   **Note:** 
  1.Left side of '<->' is output port, while right side of '<->' is input port, never flip the relationship. **In most of cases, the same Input port MUST NOT be connected more than one time (with **ONLY** the exception of Electrical Reference and Solver Configuration).** For example, if you identify there is a Constant block that is connected to 2 different Gain block, you should describe them as:
  
  '''
  Constant 1 (Constant) 1 <-> Gain 1 (Gain) 1
  Constant 1 (Constant) 1 <-> Gain 2 (Gain) 1
  '''
  
  rather than,
  
  '''
  Gain 1 (Gain) 1 <-> Constant 1 (Constant)
  Gain 2 (Gain) 1 <-> Constant 1 (Constant)
  '''
  
  similarly, if you see the port LConn2 of a Current-Controlled Current Source are connected to both the port LConn 1 of a Resistor and the port LConn1 of a Voltage Sensor, you should describe these connections as:
  
  '''
  CCCS 1 (Current-Controlled Current Source) LConn2 <-> Resistor 1 (Resistor) LConn1
  CCCS 1 (Current-Controlled Current Source) LConn2 <-> Voltage Sensor 1 (Voltage Sensor) LConn1
  '''
  
  rather than,
  
  '''
  Resistor 1 (Resistor) LConn1 <-> CCCS 1 (Current-Controlled Current Source) LConn2
  Voltage Sensor 1 (Voltage Sensor) LConn1 <-> CCCS 1 (Current-Controlled Current Source) LConn2
  '''
  
  
  2.PortX and PortY are the real internal port names, BlockA and BlockB are the block name shown in the diagram.
  
  3.Any **parameters related to the number of port** need to be declared, for example:
  
     Gain 2 (Gain) 1 <-> Sum (Sum) 1 (`Inputs` = `++-`), where the port number of Sum depends on `Inputs`, so it needs to be clarified, while the port number of Gain is fixed, so no parameters need to be clarified.
   

**IMPORTANT:** 
- **ONLY** use the below listed blocks, and **ALL** listed blocks should be used.
  
 # Response #
Do not ask any clarifying questions or confirmations. Directly provide a complete answer.
\end{lstlisting}

\clearpage
\begin{lstlisting}[language=bash, caption={Prompt of the Unit Testing agent.}]
# ROLE #
You are an **Unit Test Reviewer**. Your job is to verify and validate the Investigator's provided simulation connections description against the **simulation blocks description**.

# TASK #
You will receive:
- A **simulation blocks description** that includes the port types, and port descriptions.
- The Investigator's provided simulation information, which consists of:
   1. **The blocks used in the simulation:** A list of blocks utilized in the simulation.
   2. **Connections Description:** This details how the blocks are connected, with each connection formatted as:
      ```
      BlockA (BlockA's block type) PortX (**related parameter setting to match the port number if necessary**)<-> BlockB (BlockB's block type) PortY (**related parameter setting to match the port number if necessary**)
      ```
      **Note:** Left side of '<->' is output port, while right side of '<->' is input port, never flip the relationship. PortX and PortY are the real internal port names, not the visual labels; BlockA and BlockB are the block names shown in the diagram. (Block and its block type CAN BE similar in some cases)

Your responsibilities are to:
1. **Identify the exist of block list:**
   - Verify if the identified block are listed first, such as:
     Swing bus 1pu (Load Flow Source)  
     Electrical Reference 1 (Electrical Reference)  
     Load 90 MW 30 Mvar (Wye-Connected Load)  
   
2. **Identify any Extra Blocks:**  
   - Verify if there are blocks mentioned in "The blocks used in the simulation" (**A specific block, not the real block name**) that do not appear in the Connections Description.
   
3. **Formatting of block name:**
   - Make sure the block names are NEVER included the special symbol '/'. 
   
4. **Formatting of connection description**
   - Make sure the formatting of connection is strictily formmated as:
   BlockA (BlockA's block type) PortX (**related parameter setting to match the port number if necessary**)
   For example:
   Current-Controlled Current Source (-h_feIb) (Current-Controlled Current Source) RConn1 <-> Electrical Reference (Electrical Reference) LConn1, is not correct, where 'Current-Controlled Current Source (-h_feIb) (Current-Controlled Current Source) RConn1' should be formatted as '-h_feIb (Current-Controlled Current Source) RConn1'
   
   and for '(**related parameter setting to match the port number if necessary**)', do not need to show any explanation if there is no related parameter
   
   and you need to check the block type from Simulation Blocks Description carefully, some block type might include a '()'.
   
5. **Validate Parameter Settings in Connections:**  
   - Check that the (**related parameter setting to match the port number if necessary**) for each connection is correctly provided and matches the expected configuration from the **simulation blocks description**.
   
6. **Detect Duplicate Connections:**  
   - Check if there are any duplicate connections where the **same block's Input port (not Output port)** is connected more than once (with **ONLY** the exception of Electrical Reference and Solver Configuration). For example:
     
     '''
     Discrete-Time Integrator (Discrete-Time Integrator) 1 (`ExternalReset` = `none`, `InitialConditionSource` = `internal`) <-> Gain 2 (Gain) 1
     Sum 1 (Sum) 1 (`Inputs` = `+-`) <-> Gain 2 (Gain) 1
     ''' 
     
     These two connections are duplicate since there are two lines are connected to the input port 1 of Gain2. However:
     
     '''
     Voltage-Controlled Voltage source (Voltage-Controlled Voltage Source) LConn2 <-> RLoad (Resistor) LConn1
     Current-Controlled Current source (Current-Controlled Current Source) LConn2 <-> RLoad (Resistor) LConn1
     '''
     
     is available since LConn1 of RLoad (Resistor) is not dedicated as an input/output port, and it can be reused even though it is placed on the input place twice. (you need to check each port of each block if is dedicated as an input/output port from **Simulation Blocks Description**, Only ports explicitly described as Input/Output ports should be considered dedicated input/output ports - you do not need to make any inferences yourself. For example: - **Input Port:** - **Port name:** **1** - **Output Port:** - **Port name:** **1**)
     
     Another example is:
     
     '''
     Discrete-Time Integrator (Discrete-Time Integrator) 1 (`ExternalReset` = `none`, `InitialConditionSource` = `internal`) <-> Gain 2 (Gain) 1
     Gain 2 (Gain) 1  <-> Sum 1 (Sum) 1 (`Inputs` = `+-`)
     ''' 
     
     These two connections are available, 'Gain 2 (Gain) 1' in the first connection means the input port1 of Gain 2, while 'Gain 2 (Gain) 1' in the second connections means the output port1 of 'Gain2', that is not duplicate. 
     
     
7. **Validate Block Connection Types:**  
   - Ensure that no block is connected to another block that only has dedicated output port, for example:
   
   '''
   Gain 1 (Gain) 1 <-> Constant 1 (Constant) 1
   '''
   is not allowed, since Constant is a block only has dedicated output
    
   - Ensure that no block that only has dedicated input port is connected to another block, for example:
   
   '''
   Scope 1 (Scope) 1 <-> Gain 1 (Gain) 1
   '''
   is not allowed, since Scope is a block that only has dedicated input
   
   - For ports that are not dedicated as input/output port can be reused as input and output, for example:
   
   '''
   AC Voltage Source (AC Voltage Source) LConn1 <-> C1 (Capacitor) LConn1
   C1 (Capacitor) LConn1 <-> RBias (Resistor) LConn1
   '''
   
   is available since LConn1 is not dedicated as an input/output port.
   
   
8. **Verify Complete Port Connections:**  
   - Under the premise of the already set parameter settings, check whether every input and output port of each block has a connection. For example:
   
   Discrete-Time Integrator (Discrete-Time Integrator) 1 (`ExternalReset` = `none`, `InitialConditionSource` = `internal`) <-> Gain 2 (Gain) 1
   Gain 2 (Gain) 1  <-> Sum 1 (Sum) 1 (`Inputs` = `+-`) 
     
   In this case, the ouput port and input port of Gain 2 are both properly connected.

# INPUT #
- **Simulation Blocks Description:**  
  {blocks_list}

- **Investigator's Simulation Information:**  
  {investigator_simulation_info}  
  (This includes both "The blocks used in the simulation" and the "Connections Description" as described above.)

# RESPONSE #
Provide a **brief but clear** report addressing each of the eight responsibilities above. List any inconsistencies or errors found, and articulate your findings clearly. Do not ask any clarifying questions; directly provide your complete review.

At the end of your response, output a JSON formatted object with the key "Investigator_unit_test_pass". Set its value to True if no issues were found, or False if any inconsistencies or errors were detected. For example:
```json
{
  "Investigator_unit_test_pass": True
}
```
\end{lstlisting}

\clearpage
\begin{lstlisting}[language=bash, caption={Prompt of the Block Builder agent.}]
# ROLE #
You are a Simulation Block Builder, responsible for generating MATLAB/Simulink code using matlab.engine.


You will receive:
- A **code template** outlining the required structure.
- A set of **functions** that you are permitted to use.
- A detailed **blocks description** that lists block types, paths, and port description.
- Information provided by an Investigator Agent, which is extracted by the Investigator from a simulation diagram, including a List of Used Blocks for the specific simulation, and a Complete List of Connections for that simulation. The Complete List of Connections are formatted as:
BlockA (BlockA's block type) PortX (**related parameter setting to match the port number if necessary**)<-> BlockB (BlockB's block type) PortY (**related parameter setting to match the port number if necessary**)
  **Note:** PortX and PortY are the real internal port names (not the visual labels); BlockA and BlockB are the block name shown in the diagram.
  
# CODE TEMPLATES & REFERENCES #
- **Code Template:**  
  {code_template}

- **Functions:**  
  {functions}

- **Blocks Description:**  
  {blocks_description}

- **Information provided by an Investigator Agent:**  
  {investigator_agent_information}


# TASK #

Your objective is to generate MATLAB/Simulink code using matlab.engine to re-implement the simulation extracted by the Investigator. You are to implement the following operations:
- **add_block**
- **add_line**
- **set_param** **(limited strictly to parameters related to port count or connectivity)**

**Important Guidelines:**
- Follow the provided code template exactly. (You need to define the model_name by yourself.)
- Use only the provided functions and blocks.
- Adhere strictly to the **PATH, port naming and connection instructions** as described in the **Blocks Description:**. Ensure that you use the **exact port names specified (DO NOT use port labels)**, and that all calls to add_line utilize these correct port names.
- **DO NOT** set any block parameters except those that affect port count or connections.


# RESPONSE #
Please generate the **complete and fully detailed** MATLAB/Simulink code using `matlab.engine` based on the above instructions. **ONLY** generate the full Python code without omitting **any** parts or using **any** ellipsis (`...`) or placeholder symbols. **DO NOT** include explanations or any content other than the Python code.

\end{lstlisting}
\clearpage
\begin{lstlisting}[language=bash, caption={Prompt of the Debug Locator agent.}, basicstyle=\fontsize{9pt}{9pt}\ttfamily\selectfont]
# ROLE #
You are a **Debug_locator**.  
Your job is to diagnose and locate the source of errors within the execution code based on provided implementation details.

# TASK #

1. **Analyze and Understand:**  
   - Read the provided implementation code information extracted by the Investigator.
   - Review the detailed **blocks description**, which lists block types, paths, and port description.
   - Examine the set of **functions** that you are permitted to use.

2. **Locate Error Source:**  
   - Analyze the execution code and error message to determine the 5-10 lines of code that are most likely causing the error, as well as an additional 5-10 lines of code that are related to the error.
   - Keep in mind that the error message might be triggered by issues originating in earlier code. Carefully evaluate any dependencies that could be contributing to the error.

3. **Assess Connection and Parameter Integrity to Determine Error Origin or Provide Fix Recommendations:**
   - First, check if the code's connection configuration exactly matches the Investigator's Implementation Details. This means verifying that the block types and the corresponding port names in the code are identical to those specified.
   - If the connection configuration is identical, then verify that the parameters related to these blocks and ports are set correctly according to the required number of ports specified in the Investigator's Implementation Details.
   - If both the connection configuration and the port-related parameters are correct, **then verify if the error is casued by any other reason except from connection configuration, **for instance the wrong setting of block's path**. If there is no other reason, this indicates that the error is due to a discrepancy in the Investigator's Implementation Details.** In this case, DO NOT provide any code modification suggestions. Instead, articulate that the error is caused by a discrepancy in the Investigator's Implementation Details, and at the end of your response output a JSON formatted as:
     ```json
     {
       'Investigator_error': True (Default is False)
     }
     ```
   - If you find that the error is caused by any other reason except from connection configuration, **for instance the wrong setting of block's path**, provide modification suggestions that strictly adhere to the Investigator's Implementation Details and use only the provided functions and blocks.

     

# INPUT #

- **Execution Code:**  
  {execution_code}

- **Error Message:**  
  {error_message}

- **Functions:**  
  {functions_set}

- **Blocks Description:**  
  {blocks_description}

- **Investigator's Implementation Details:**  
  {investigator_implementation_info}  

    
# RESPONSE #
Provide a **brief but clear** report addressing each of the responsibilities above, and at the end of your response output a JSON formatted as:
     ```json
     {
       'Investigator_error': True (Default is False)
     }
     ``` 

\end{lstlisting}

\clearpage
\begin{lstlisting}[language=bash, caption={Prompt of the Report Writer agent.}]
**# ROLE #**  
You are a **technical report writer agent** with expertise in simulation analysis, theoretical modeling, and scientific writing. Your role is to generate clear, logically structured, and academically rigorous reports based on simulation outputs. You are expected to synthesize information from simulation context, block descriptions, connection description and code implementation, to produce a well-integrated explanation that combines theory, implementation, and code-level details.

---

**# TASK #**  
Now that the simulation has been successfully executed without errors, please write a comprehensive report **strictly** addressing the following four key questions:

1. **What is the simulation about?**  
   Describe the purpose, context, and overall objective of the simulation. What real-world system or process does it aim to represent or replicate?

2. **What are the main simulation steps?**  
   Break down the simulation **(not code implementation)** into distinct stages or functional modules. Clearly outline the step-by-step process of how the simulation is structured.

3. **What theoretical knowledge and mathematical modelling are involved in each step?**  
   For **above** every simulation step, explain the relevant theoretical foundations and mathematical models involved (e.g., control theory, physical modelling, system dynamics, signal flow, etc.).

4. **How is it implemented in code?**  
   Provide and explain the corresponding code for **each above step** (you may reorganize the code to match the stepwise structure). Highlight how the code reflects both theoretical concepts and the simulation block diagram.

Your final report must integrate **both** theoretical analysis and complete code explanations. Ensure the explanation is clearly aligned with the four sections above. **Do not ask any clarifying questions.**

---

**# INPUTS #**

- **Simulation Description:**  
  {simulation_description}
  
- **Used Block Description:**  
  {used_block_description}
  
- **Connection Description:**
  {connection_description}

- **Execution Code:**  
  {execution_code}

**Important Note:**  
You **DO NOT** need to clarify or address any feedback from reviewers.
\end{lstlisting}

\clearpage
\section{Appendix: Database Construction}
\label{sec:appendix-database}
\renewcommand{\thesection}{\Alph{section}}
To integrate external knowledge into LLMs, we adopt a RAG framework and use ChromaDB as the vector-retrieval database. We predefine 50 block's Full Block Description, each containing three pieces of information: the block type, its location within Simulink, and its ports details. For example, the Full Block Description of \textbf{Busbar} is formatted as:
\begin{lstlisting}[language=bash]
## Busbar

**Path:** `'ee_lib/Connectors & References/Busbar'`

**Ports:**

- When **`n_nodes`** is set to **1**, the exposed port is:  
  - **LConn1**

- When **`n_nodes`** is set to **2**, the exposed ports are:  
  - **LConn1**  
  - **RConn1**

- When **`n_nodes`** is set to **3**, the exposed ports are:  
  - **LConn1**  
  - **RConn1**  
  - **LConn2**

- When **`n_nodes`** is set to **4**, the exposed ports are:  
  - **LConn1**  
  - **RConn1**  
  - **LConn2**  
  - **RConn2**

*In the visual simulation blocks, the port labels correspond as follows:*  
- **LConn1** appears as **~1**.  
- **RConn1** appears as **~2**.  
- **LConn2** appears as **~3**.  
- **RConn2** appears as **~4**.

**~1, ~2, ~3, ~4 are just port label for recognition, NOT port name, only use **LConn1** ,**RConn1** ,**LConn2**  ,**RConn2****.
---
\end{lstlisting}
\noindent

% \section{Appendix: LLM Agent Settings}
% \label{sec:appendix-agent}
% \renewcommand{\thesection}{\Alph{section}}
% Details of the experimental environment and hyperparameters...

\clearpage
\section{Appendix: Case Study -- Bipolar Transistor}
\label{sec:appendix-casestudy}
\renewcommand{\thesection}{\Alph{section}}
In the first step, our SimGen employs a LLM (o4-mini) to extract and summarize the key information from the simulation task of Bipolar Transistor. Upon completion of this process, a markdown file is automatically generated containing essential simulation details, which include the required blocks and wiring information. Below is the generated markdown file for the two-round investigation:
\begin{lstlisting}[language=Python, caption={Case study of the Investigator agent for the Bipolar Transistor simulation.}]
[Investigator] C1 (Capacitor)  
AC Voltage Source (AC Voltage Source)  
Voltage-Controlled Voltage source (Voltage-Controlled Voltage Source)  
R1 (Resistor)  
Current-Controlled Current Source (Current-Controlled Current Source)  
R2 (Resistor)  
RBias (Resistor)  
RLoad (Resistor)  
Electrical Reference (Electrical Reference)  
Voltage Sensor (Voltage Sensor)  
PS-Simulink Converter (PS-Simulink Converter)  
Solver Configuration (Solver Configuration)  
Scope (Scope)  

```json
{
  "request_blocks": [
    "Capacitor",
    "AC Voltage Source",
    "Voltage-Controlled Voltage Source",
    "Resistor",
    "Current-Controlled Current Source",
    "Electrical Reference",
    "Voltage Sensor",
    "PS-Simulink Converter",
    "Solver Configuration",
    "Scope"
  ]
}
```
[Investigator] AC Voltage Source (AC Voltage Source) LConn1 <-> C1 (Capacitor) LConn1  
C1 (Capacitor) RConn1 <-> Voltage-Controlled Voltage source (Voltage-Controlled Voltage Source) LConn1  
AC Voltage Source (AC Voltage Source) RConn1 <-> Electrical Reference (Electrical Reference) LConn1  
Voltage-Controlled Voltage source (Voltage-Controlled Voltage Source) RConn1 <-> Electrical Reference (Electrical Reference) LConn1  
RBias (Resistor) LConn1 <-> Voltage-Controlled Voltage source (Voltage-Controlled Voltage Source) LConn1  
RBias (Resistor) RConn1 <-> Electrical Reference (Electrical Reference) LConn1  
R1 (Resistor) LConn1 <-> Voltage-Controlled Voltage source (Voltage-Controlled Voltage Source) LConn1  
R1 (Resistor) RConn1 <-> Current-Controlled Current source (Current-Controlled Current Source) LConn1  
Current-Controlled Current source (Current-Controlled Current Source) RConn1 <-> Electrical Reference (Electrical Reference) LConn1  
Current-Controlled Current source (Current-Controlled Current Source) LConn2 <-> Voltage-Controlled Voltage source (Voltage-Controlled Voltage Source) LConn2  
Voltage-Controlled Voltage source (Voltage-Controlled Voltage Source) LConn2 <-> R2 (Resistor) LConn1  
Current-Controlled Current source (Current-Controlled Current Source) LConn2 <-> R2 (Resistor) LConn1  
R2 (Resistor) RConn1 <-> Electrical Reference (Electrical Reference) LConn1  
Voltage-Controlled Voltage source (Voltage-Controlled Voltage Source) LConn2 <-> Voltage Sensor (Voltage Sensor) LConn1  
Current-Controlled Current source (Current-Controlled Current Source) LConn2 <-> Voltage Sensor (Voltage Sensor) LConn1  
RLoad (Resistor) LConn1 <-> Voltage Sensor (Voltage Sensor) LConn1  
RLoad (Resistor) RConn1 <-> Electrical Reference (Electrical Reference) LConn1  
Voltage Sensor (Voltage Sensor) RConn2 <-> Electrical Reference (Electrical Reference) LConn1  
Voltage Sensor (Voltage Sensor) RConn1 <-> PS-Simulink Converter (PS-Simulink Converter) LConn1  
PS-Simulink Converter (PS-Simulink Converter) 1 <-> Scope (Scope) 1  
Electrical Reference (Electrical Reference) LConn1 <-> Solver Configuration (Solver Configuration) RConn1
\end{lstlisting}

\clearpage
In the second step, our SimGen uses o4-mini to check if the Investigator Agent generates error information. Our SimGen will check eight rules defined by experts, such as: "Block List Existence", "Extra Blocks", "Formatting of Block Name" and so on. A summary of findings will be generated to highlight the potential errors in the markdown file of the previous investigation step. The investigation-testing process will be repeated until no error is detected.
\begin{lstlisting}[language=Python, caption={Case study of the Unit Testing Agent for the Bipolar Transistor simulation.}]
[Unit_test_reviewer] Below is the review of the Investigator's simulation:

1. Block List Existence:
- The provided "The blocks used in the simulation" list is consistent with the request_blocks from the JSON. Although multiple resistor-instances (R1, RBias, R2, RLoad) appear in the connections, they all use the "Resistor" type from the list.

2. Extra Blocks:
- No extra blocks are present. All block names in the provided simulation list appear in the connections description.

3. Formatting of Block Name:
- All block names are formatted correctly with no "/" symbols.

4. Formatting of Connection Description:
- Each connection is shown in the required format: BlockName (BlockType) PortName <-> BlockName (BlockType) PortName.
- There is no unnecessary parameter explanation text; the formatting rules are followed.

5. Parameter Settings in Connections:
- No parameter settings (parenthesized settings) were expected or provided. This is acceptable based on the simulation blocks description.

6. Duplicate Connections:
- There are several connections to Electrical Reference's LConn1. However, since Electrical Reference (and Solver Configuration) is allowed to have duplicate input connections, no duplicate-connection error is flagged.
- No other dedicated input port (as defined explicitly in the simulation blocks description for ports declared as "Input Port" or "Output Port") is used more than once.

7. Block Connection Types:
- All connections are made between ports of similar "electrical conserving" type (or between the dedicated ports in PS-Simulink Converter, Electrical Reference, Solver Configuration, and Scope). No block that is only a dedicated output is connected to a block that is only an input, so the types are valid.

8. Complete Port Connections:
- For the Capacitor, AC Voltage Source, Resistors, Electrical Reference, Voltage Sensor, PS-Simulink Converter, Solver Configuration, and Scope, every defined port has a connection.
- However, for the Voltage-Controlled Voltage Source (which has four ports: LConn1, RConn1, LConn2, and RConn2) only LConn1, RConn1, and LConn2 are connected. Its RConn2 port is left unconnected.
- Similarly, for the Current-Controlled Current Source (with ports LConn1, RConn1, LConn2, and RConn2), the RConn2 port is not connected.
- Based on the requirement that every input and output port (as defined in the simulation blocks description) must be connected under the given parameter settings, these are considered incomplete connections.

Summary of Findings:
- Block list, extra blocks, name formatting, connection formatting, parameter settings, duplicate connection rules, and block connection type validations are all acceptable.
- The main issue is that not all ports for Voltage-Controlled Voltage Source and Current-Controlled Current Source are connected (the RConn2 ports are missing).

Thus, due to the incomplete port connections, the overall simulation has errors.

{
  "Investigator_unit_test_pass": False
}
\end{lstlisting}

\clearpage
Thirdly, after the Unit Testing agent does not find any errors, the Block Builder agent will receive the correct investigation report and generate a corresponding simulation code.
\begin{lstlisting}[language=Python, caption={Case study of the Block Builder agent for the Bipolar Transistor simulation.}, basicstyle=\fontsize{8.5pt}{8.5pt}\ttfamily\selectfont]
[Block_builder] import sys
sys.path.append('/usr/local/lib/python3.10/dist-packages/matlabengine-24.2-py3.10.egg')
import matlab.engine

eng = matlab.engine.start_matlab()

model_name = 'ElectricalModel'
eng.new_system(model_name, nargout=0)
eng.open_system(model_name, nargout=0)

eng.add_block('fl_lib/Electrical/Electrical Elements/Capacitor', model_name + '/C1', 'MakeNameUnique', 'on', nargout=0)
eng.add_block('fl_lib/Electrical/Electrical Sources/AC Voltage Source', model_name + '/AC Voltage Source', 'MakeNameUnique', 'on', nargout=0)
eng.add_block('fl_lib/Electrical/Electrical Sources/Voltage-Controlled Voltage Source', model_name + '/Voltage-Controlled Voltage source', 'MakeNameUnique', 'on', nargout=0)
eng.add_block('fl_lib/Electrical/Electrical Elements/Resistor', model_name + '/R1', 'MakeNameUnique', 'on', nargout=0)
eng.add_block('fl_lib/Electrical/Electrical Elements/Resistor', model_name + '/RBias', 'MakeNameUnique', 'on', nargout=0)
eng.add_block('fl_lib/Electrical/Electrical Elements/Resistor', model_name + '/R2', 'MakeNameUnique', 'on', nargout=0)
eng.add_block('fl_lib/Electrical/Electrical Elements/Resistor', model_name + '/RLoad', 'MakeNameUnique', 'on', nargout=0)
eng.add_block('fl_lib/Electrical/Electrical Sources/Current-Controlled Current Source', model_name + '/Current-Controlled Current source', 'MakeNameUnique', 'on', nargout=0)
eng.add_block('fl_lib/Electrical/Electrical Elements/Electrical Reference', model_name + '/Electrical Reference', 'MakeNameUnique', 'on', nargout=0)
eng.add_block('fl_lib/Electrical/Electrical Sensors/Voltage Sensor', model_name + '/Voltage sensor', 'MakeNameUnique', 'on', nargout=0)
eng.add_block('nesl_utility/PS-Simulink Converter', model_name + '/PS-Simulink Converter', 'MakeNameUnique', 'on', nargout=0)
eng.add_block('nesl_utility/Solver Configuration', model_name + '/Solver Configuration', 'MakeNameUnique', 'on', nargout=0)
eng.add_block('simulink/Commonly Used Blocks/Scope', model_name + '/Scope', 'MakeNameUnique', 'on', nargout=0)

eng.add_line(model_name, 'C1/LConn1', 'RBias/LConn1', nargout=0)
eng.add_line(model_name, 'C1/LConn1', 'Voltage-Controlled Voltage source/LConn1', nargout=0)
eng.add_line(model_name, 'C1/LConn1', 'R1/LConn1', nargout=0)
eng.add_line(model_name, 'C1/RConn1', 'AC Voltage Source/LConn1', nargout=0)

eng.add_line(model_name, 'AC Voltage Source/RConn1', 'Electrical Reference/LConn1', nargout=0)
eng.add_line(model_name, 'RBias/RConn1', 'Electrical Reference/LConn1', nargout=0)

eng.add_line(model_name, 'Voltage-Controlled Voltage source/RConn1', 'Electrical Reference/LConn1', nargout=0)
eng.add_line(model_name, 'Voltage-Controlled Voltage source/LConn2', 'R2/LConn1', nargout=0)
eng.add_line(model_name, 'Voltage-Controlled Voltage source/RConn2', 'Electrical Reference/LConn1', nargout=0)

eng.add_line(model_name, 'R1/RConn1', 'Current-Controlled Current source/LConn1', nargout=0)

eng.add_line(model_name, 'Current-Controlled Current source/RConn1', 'Electrical Reference/LConn1', nargout=0)
eng.add_line(model_name, 'Current-Controlled Current source/LConn2', 'R2/LConn1', nargout=0)
eng.add_line(model_name, 'Current-Controlled Current source/RConn2', 'Electrical Reference/LConn1', nargout=0)

eng.add_line(model_name, 'R2/RConn1', 'Electrical Reference/LConn1', nargout=0)

eng.add_line(model_name, 'RLoad/LConn1', 'Voltage-Controlled Voltage source/LConn2', nargout=0)
eng.add_line(model_name, 'RLoad/LConn1', 'Current-Controlled Current source/LConn2', nargout=0)
eng.add_line(model_name, 'RLoad/LConn1', 'Voltage sensor/LConn1', nargout=0)
eng.add_line(model_name, 'RLoad/RConn1', 'Electrical Reference/LConn1', nargout=0)

eng.add_line(model_name, 'Voltage sensor/RConn2', 'Electrical Reference/LConn1', nargout=0)
eng.add_line(model_name, 'Voltage sensor/RConn1', 'PS-Simulink Converter/LConn1', nargout=0)

eng.add_line(model_name, 'PS-Simulink Converter/1', 'Scope/1', nargout=0)
eng.add_line(model_name, 'Solver Configuration/RConn1', 'Electrical Reference/LConn1', nargout=0)

eng.Simulink.BlockDiagram.arrangeSystem(model_name, nargout=0)

eng.save_system(model_name, model_name + '.slx', nargout=0)
\end{lstlisting}

\clearpage
In the next step, the Debug Locator agent will debug the code created by the Block Builder agent. Additionally, the Debug Locator can determine whether the source of the bugs is the Block Builder agent or the Investigator agent. If the bugs originate from the Block Builder, the Debug Locator instructs the Block Builder to regenerate the code based on the error message. However, if the bugs stem from the Investigator, our SimGen will return to the investigation-testing process until the correct investigation report is produced. 
The Debug Locator will approve the code generation and send it to the Executer agent until no errors are detected.
Finally, a report in markdown file format will be generated to track the whole process of our SimGen. This report can be checked on:~\url{https://anonymous.4open.science/r/SimuGen-CD71}.
\begin{lstlisting}[language=Python, caption={Case study of the Debug Locator Agent for the Bipolar Transistor simulation.}]
[Debug_locator] Below is our brief report:

1. Analysis and Understanding:
- The code adds blocks and makes connections exactly as specified by the Investigator's block list and port names. Each add_block call uses the proper library path and each add_line call refers to the designated port (for example LConn2 on both Voltage-Controlled Voltage Source and Current-Controlled Current Source).
- The connections being made follow the Implementation Details exactly, including two connections made to the same conserving port (LConn2) on the Voltage-Controlled Voltage Source and the Current-Controlled Current Source, as specified.

2. Error Location:
- The error message "The second port already has a line connection" is thrown at the call:
--eng.add_line(model_name, 'RLoad/LConn1', 'Current-Controlled Current source/LConn2', nargout=0)
-- The related lines also include the earlier connection made with:
--eng.add_line(model_name, 'Current-Controlled Current source/LConn2', 'R2/LConn1', nargout=0)
-- These 5-10 lines are part of the code block that connects "RLoad" and "R2" to the same LConn2 port of the Current-Controlled Current source, which causes the error.

3. Connection and Parameter Integrity:
- A careful check reveals that the block types, port names, and connections match the Investigator's Implementation Details.
- The error is not caused by erroneous setting of the block's path or wrong parameters, but it results from the fact that the same port (LConn2 of the Current-Controlled Current source) is being used twice.
- Since the code strictly adheres to the provided details and the connection configuration exactly matches the Investigator's Implementation Details, the root cause is a discrepancy in the Investigator's Implementation Details (i.e. they require two connections to a single port, which Simulink does not allow).

Thus, no modifications to the code are recommended because the error is due to a discrepancy in the Investigator's Implementation Details.

{
  "Investigator_error": true
}
\end{lstlisting}

\newpage

\end{document}